\documentclass[preprint,11pt]{elsarticle}
\usepackage{geometry}
\usepackage{amssymb}
\usepackage{subfigure}
\usepackage{longtable}
\usepackage{multirow}
\usepackage{array}
\usepackage{color}
\usepackage[ruled]{algorithm2e}
\usepackage{float}
\biboptions{numbers,sort&compress}
\usepackage[numbers,sort&compress]{natbib}
\usepackage{amsmath}
\usepackage{diagbox}
\usepackage{hyperref}
\usepackage{makecell}
\UseRawInputEncoding

\journal{Information Sciences}

\begin{document}

\begin{frontmatter}

\title{FGTBT: Frequency-Guided Task-Balancing Transformer for Unified Facial Landmark Detection}

\author[firstauthor2,firstauthor]{Jun Wan\fnref{equalcontrib}}
\author[firstauthor2,firstauthor]{Xinyu Xiong\fnref{equalcontrib}}
\author[firstauthor2]{Ning Chen\corref{cor2}}
\author[thirdauthor]{Zhihui Lai}
\author[thirdauthor]{Jie Zhou}
\author[fourthauthor]{Wenwen Min}

\address[firstauthor2]{School of Information Engineering, Zhongnan University of Economics and Law, Wuhan, China}

\address[firstauthor]{School of Computer Science and Engineering, Nanyang Technological University, Singapore}

\address[thirdauthor]{College of Computer Science and Software Engineering, Shenzhen University, Shenzhen, China}

\address[fourthauthor]{School of Information Science and Engineering, Yunnan University, Kunming, Yunnan, China}

\cortext[cor2]{Corresponding author: Ning Chen (safetychn@zuel.edu.cn.)}

\fntext[equalcontrib]{These authors contributed equally to this work.}

\begin{abstract}
Recently, deep learning based facial landmark detection (FLD) methods have achieved considerable success. However, in challenging scenarios such as large pose variations, illumination changes, and facial expression variations, they still struggle to accurately capture the geometric structure of the face, resulting in performance degradation. Moreover, the limited size and diversity of existing FLD datasets hinder robust model training, leading to reduced detection accuracy. To address these challenges, we propose a Frequency-Guided Task-Balancing Transformer (FGTBT), which enhances facial structure perception through frequency-domain modeling and multi-dataset unified training. Specifically, we propose a novel Fine-Grained Multi-Task Balancing loss (FMB-loss), which moves beyond coarse task-level balancing by assigning weights to individual landmarks based on their occurrence across datasets. This enables more effective unified training and mitigates the issue of inconsistent gradient magnitudes. Additionally, a Frequency-Guided Structure-Aware (FGSA) model is designed to utilize frequency-guided structure injection and regularization to help learn facial structure constraints. Extensive experimental results on popular benchmark datasets demonstrate that the integration of the proposed FMB-loss and FGSA model into our FGTBT framework achieves performance comparable to state-of-the-art methods. The code is available at \href{https://github.com/Xi0ngxinyu/FGTBT}{https://github.com/Xi0ngxinyu/FGTBT}.
\end{abstract}

\begin{keyword}
facial landmark detection, heatmap regression, transformer, large pose, facial expression.
\end{keyword}

\end{frontmatter}


\section{Introduction}
\label{}

Facial landmark detection (FLD), also known as face alignment, aims to locate a set of landmarks on the face with special semantic information (such as eye corners, facial outline, bridge of the nose and so on). It is a valuable and fundamental research direction for many face analysis tasks, such as facial expression recognition\cite{liu2025facial}, face recognition\cite{LIU2024120470} and driver fatigue recognition\cite{LI2023833}. 

FLD has advanced significantly with the development of convolutional neural networks (CNNs)\cite{gao2023ctcnet}, and primarily follows two paradigms: coordinate regression and heatmap regression. Coordinate regression methods \cite{sun2013deep,  trigeorgis2016mnemonic} directly predict landmark coordinates and they are computationally efficient and easy to implement, making them well-suited for real-time applications. In contrast, heatmap regression methods \cite{kowalski2017deep, yang2017stacked, yue2021multi} generate a probability heatmap for each landmark, offering superior performance due to their ability to preserve spatial consistency and model complex facial structures. Recently, Transformers\cite{xia2022sparse} have also been introduced into facial landmark detection tasks. With their strong global modeling capability, they demonstrate great potential in capturing long-range dependencies and complex facial structures. However, due to the high computational cost and the tendency to overfit on small-scale datasets, their practical application still faces a trade-off between efficiency and generalization. Although heatmap regression and Transformer-based methods have advanced the development of face alignment, their performance remains unsatisfactory in challenging scenarios such as large pose variations or severe illumination changes. These difficulties arise primarily from two issues:
(1) The limited availability of high-quality annotated data. Since manually labeling facial landmarks is both labor-intensive and time-consuming, the resulting data scarcity hinders the model's ability to learn robust and generalizable representations;
(2) Facial structures are challenging to model accurately under large pose variations, facial expressions, and illumination changes, which further leads to degraded detection performance.
 
\begin{figure}[!t]
\centering
\includegraphics[width=0.7\linewidth]{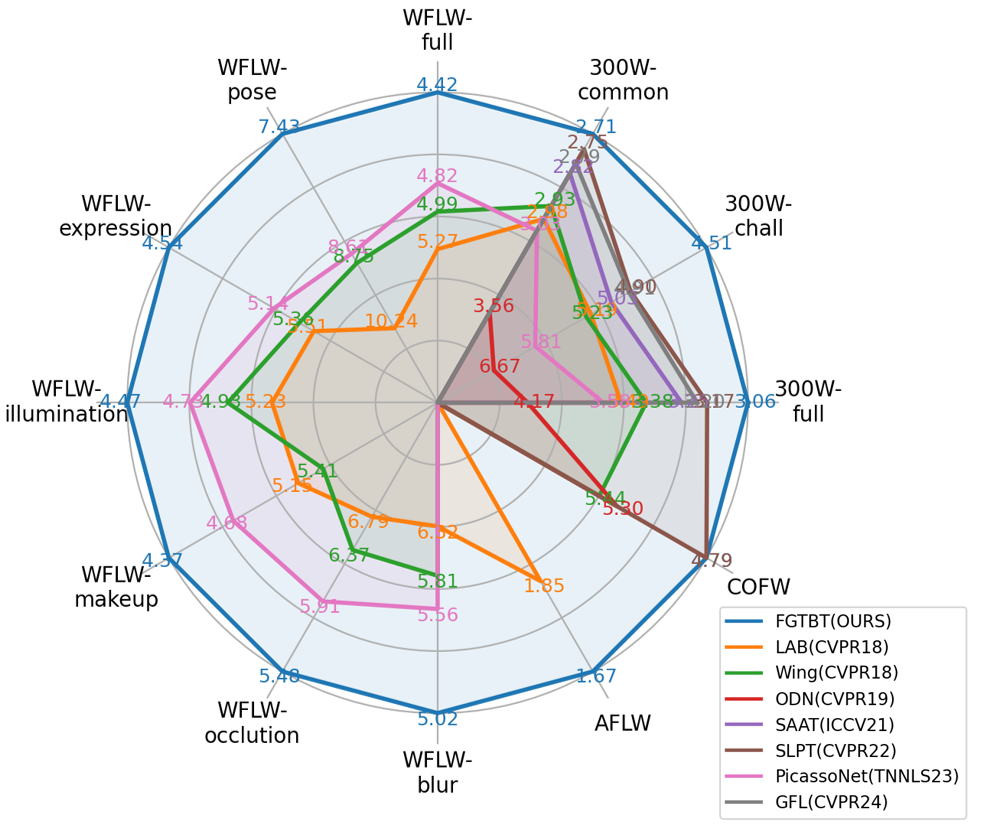}
\centering
\caption{The proposed FGTBT enables unified training across multiple facial landmark detection datasets with different annotation protocols and achieves superior performance.}
\label{problem}
\vspace{-1em}
\end{figure}

By analyzing existing FLD datasets, we observe that they typically share a common subset of landmarks. Leveraging this overlap through joint training can enhance detection accuracy across datasets. In addition, the dataset-specific landmarks can provide complementary structural cues, enriching the overall representation of facial structure. These observations motivate the construction of a unified FLD framework that jointly trains on multiple datasets annotated with different landmark protocols (as shown in Fig. \ref{problem}). This paradigm closely aligns with the philosophy of multi-task learning (MTL)\cite{chen2022jspnet}, where related tasks are optimized collaboratively via shared representations to improve generalization. However, most existing multi-task learning (MTL) methods \cite{chen2018gradnorm} adopt coarse-grained balancing strategies at the task level, often overlooking the finer-grained relationships between tasks. As a result, directly applying task-level balancing strategies to FLD may lead to gradient conflicts due to imbalanced landmark supervision across datasets, which can hinder model optimization and ultimately limit the accuracy of landmark detection.

To address these issues, we propose frequency-guided task-balancing transformer (FGTBT) for unified FLD. FGTBT consists of two main components: the Fine-grained Multi-task Balancing loss (FMB-loss) and the Frequency-Guided Structure-Aware (FGSA) model. FMB-loss is designed to facilitate balanced optimization across multiple datasets with heterogeneous landmark annotation protocols. By adaptively weighting individual landmarks based on their frequency and significance, the model effectively integrates information from diverse data sources and mitigates supervision imbalance. The FGSA model is proposed to introduce high-frequency information to model facial structural constraint with frequency-guided structure injection and regularization. By incorporating the FMB-loss and FGSA model within our proposed FGTBT, more robust and accurate unified FLD can be achieved. The main contributions of this work are summarized as follows:

1) A novel fine-grained multi-task balancing loss (FMB-loss) is designed to achieve unified facial landmark detection across datasets with different annotation protocols.  FMB-loss provides fine-grained landmark-level supervision signals, alleviating gradient magnitude inconsistency during optimization and thereby ensuring more accurate unified FLD.

2) By introducing frequency-guided injection and regularization, the proposed Frequency-Guided Structure-Aware (FGSA) model effectively exploits geometric structural cues embedded in high-frequency information. This, in turn, enhances the model's robustness and performance under challenging conditions, including expression variations, large pose, and illumination changes.

3) We propose a novel model, FGTBT, which integrates FGSA model and FMB-loss to address the challenges of unified facial landmark detection. Our FGTBT achieves competitive performance compared to state-of-the-art methods on benchmark datasets such as AFLW \cite{koestinger2011annotated}, WFLW \cite{wu2018look}, COFW \cite{burgos2013robust}, 300W \cite{sagonas2016300}.

The remainder of this paper is structured as follows. Section \textbf{II} provides an overview of related work. Section \textbf{III} presents the proposed approach, detailing the FMB-loss and FGSA model. In Section \textbf{IV}, we conduct a series of experiments to evaluate the effectiveness of our method. Finally, Section \textbf{V} concludes the paper.

\section{Related Work}
\indent In this section, we will introduce the recent works in facial landmark detection, multi-task and multi-objective optimization.
\subsection{Facial landmark detection}
\indent Facial landmark detection has been studied since the 1990s, with early approaches primarily based on statistical shape and appearance modeling, which relied on handcrafted representations and strong parametric assumptions to capture facial geometry and texture variations. Nowadays, deep learning based approaches have significantly improved FLD accuracy, evolving into two main branches: coordinate regression-based and heatmap regression-based methods.

\textbf{Coordinate regression-based methods.} These methods directly predict facial landmark coordinates by minimizing the error between estimated and ground truth positions. Sun et al.\cite{sun2013deep} pioneer the use of deep convolutional neural networks (CNNs) for facial landmark detection, introducing a cascaded regression approach that significantly improved accuracy. To further refine predictions, MDM\cite{trigeorgis2016mnemonic} integrates recurrent refinement mechanisms, enabling sequential adjustments for more precise localization. DeCaFA\cite{dapogny2019decafa} employs a fully convolutional U-Net to maintain spatial resolution throughout the regression process, enhancing face alignment accuracy. LAB\cite{wu2018look} incorporates facial boundary heatmaps to guide the learning process, reinforcing shape constraints for more robust predictions. More recently, DTLD \cite{li2022towards} leverages a Transformer-based decoder to extract multi-level image features, improving the model's ability to capture global and local relationships. DSLPT\cite{xia2023robust} explores landmark representations from local image patches, utilizing relation layers to model dependencies between facial landmarks. Liang et al.\cite{liang2024generalizable} propose a conditional face warping framework that uses real faces and unlabeled stylized faces to train a generalizable facial landmark detector.

\textbf{Heatmap regression-based methods.} Heatmap regression-based methods have become a dominant approach in facial landmark detection due to their ability to preserve spatial information. These methods generate probability heatmaps for each landmark and identify the landmark with the highest intensity as the final prediction. Early works, such as Yang et al.\cite{yang2017stacked}, introduce a stacked hourglass network that captures multi-scale features through repeated downsampling and upsampling operations, significantly enhancing landmark localization performance. Wang et al.\cite{wang2019adaptive} propose AWing Loss to address the imbalance between foreground and background pixels. Kumar et al.\cite{kumar2020luvli} introduce a novel framework that jointly predicts facial landmark locations, their associated uncertainties and visibility using a deep network trained with the LUVLi loss. To tackle error-bias in face alignment, Huang et al.\cite{huang2021adnet} propose ADNet, an end-to-end training pipeline that integrates anisotropic direction loss and an anisotropic attention module. Moreover, STARLoss\cite{zhou2023star} is introduced to address the semantic ambiguity problem, further improving facial landmark detection accuracy. Additionally, DSAT\cite{wan2024precise} introduces a dynamic framework that adaptively learns specialized features, significantly improving face alignment performance under challenging conditions.

\subsection{Multi-task and multi-objective optimization}
\indent Multi-task and multi-objective optimization aims to optimize multiple tasks within a model simultaneously. Balancing task losses is a key challenge in the process of optimization, as different tasks often have varying convergence rates, differing levels of importance and negative transfer issues. Numerous methods have been proposed to address these challenges. Kendall et al. \cite{kendall2018multi} introduce a method to handle homoscedastic uncertainty across tasks by using uncertainty weighting. Sener et al. \cite{sener2018multi} apply the Pareto optimal approach to optimize tasks without influencing the performance of others. What's more, some studies \cite{navon2020auxiliary} focus on minimizing the interference of auxiliary tasks to better preserve the performance of the main task. To address the issue of varying convergence rates, DWA \cite{liu2019end} and Harmony \cite{wu2024harmony} dynamically adjust task weights based on loss change rates. Additionally, PcGrad\cite{yu2020gradient} and GradNorm\cite{chen2018gradnorm} have tackled negative transfer issues from a gradient-based perspective, achieving improvements in MTL performance. Moreover, GRIDS\cite{cao2024grids} groups similar tasks enabling efficient multi-degradation restoration with adaptive model selection.

Although current FLD methods have achieved promising results, they still struggle with challenges such as inconsistent landmark definitions and imbalanced attention across datasets when trained jointly on multiple datasets. To address this issue, we introduce a Fine-grained Multi-task Balancing loss (FMB-loss), which enables unified training across datasets by adaptively balancing the contribution of each dataset from a novel perspective. Furthermore,  we also propose the Frequency-Guided Structure-Aware(FGSA) model, which enhances the model's structural perception and robustness for more accurate landmark detection.

\begin{figure*}[!t]
	\centering
	\includegraphics[width=\linewidth]{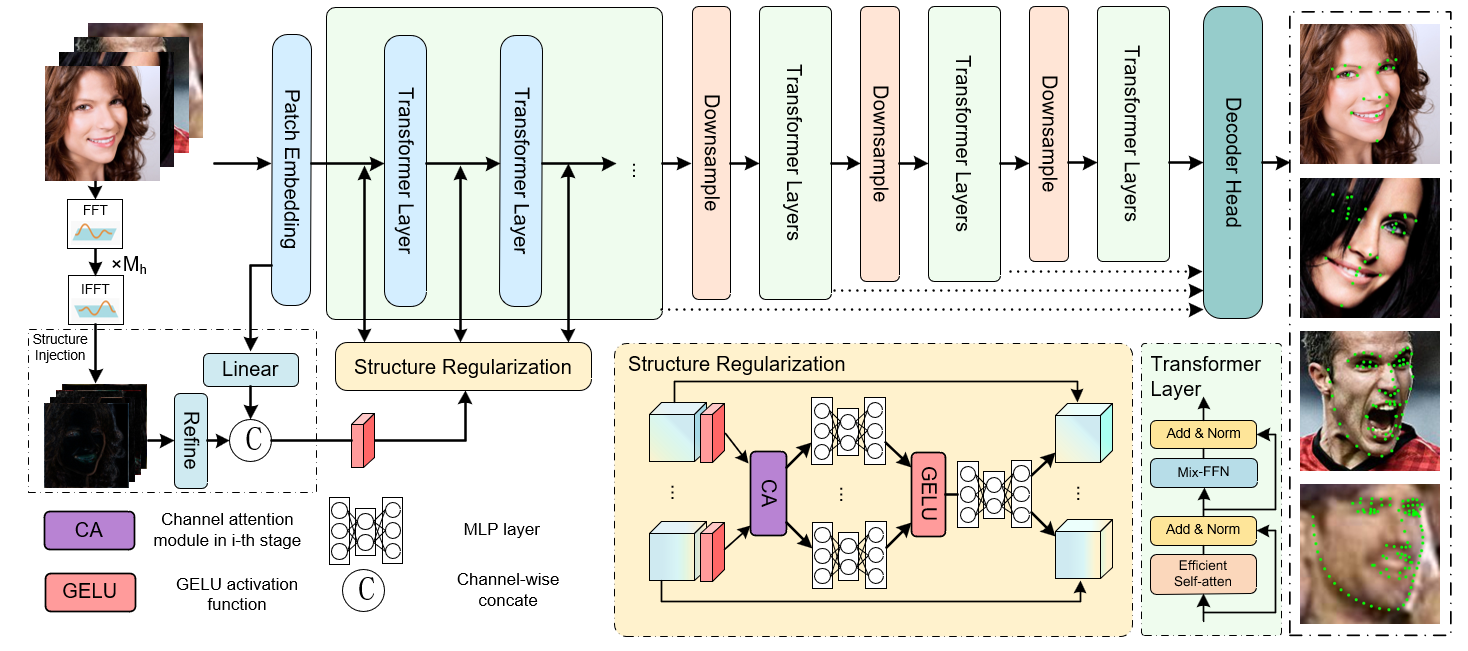}
	\centering
	\caption{The overall architecture of the proposed FGTBT. The input image is first processed through a Patch Embedding layer, followed by four hierarchical Transformer stages connected via downsampling. At each stage, frequency-domain information is first utilized to generate structure features, which are then effectively integrated into the original model using the structure injection and regularization. The features at each stage are finally decoded into facial landmarks with a decoder head. By explicitly incorporating frequency-domain structure guidance, FGTBT significantly improves structural awareness and face alignment accuracy.}
	\label{fig2}
\end{figure*}

\section{Frequency-Guided Task-Balancing Transformer for Unified Facial Landmark Detection}
In this section, we present our proposed Frequency-Guided Task-Balancing Transformer (FGTBT) in three parts. \textbf{Section III. A} introduces the overall architecture of FGTBT \textbf{Section III. B} discusses the fine-grained multi-task balancing loss, while \textbf{Section III. C} introduces the frequency-guided structure-aware model.
\subsection{Overall architecture of the proposed FGTBT}
As shown in Fig.\ref{fig2}, the proposed FGTBT adopts a coarse-to-fine architecture designed to enhance facial landmark detection under complex conditions such as pose variations, facial expressions and occlusions. The framework effectively integrates frequency-domain boundary cues and multi-dataset information into a unified model, improving structural perception and cross-dataset generalization.

The FGTBT encoder follows a hierarchical structure inspired by SegFormer\cite{xie2021segformer}. The input image $I \in \mathbb{R}^{H \times W \times 3}$ first undergoes patch embedding and is then processed through four successive stages, each containing multiple Transformer layers, respectively. The spatial resolution decreases progressively across stages, each Transformer layer incorporates efficient self-attention mechanisms and Mix-FFN modules \cite{xie2021segformer}. Moreover, in order to enhance the model's perception of facial structure, a Frequency-Guided Boundary-Aware model is integrated into each Transformer layer to strengthen the facial shape constraints. Details can be found in \textbf{Section III. C}.

In the decoding phase, multi-scale features from the four stages are upsampled to a unified spatial resolution and then concatenated. After that, they are passed through convolutional layers to generate facial landmark heatmaps. Since the proposed FGTBT is trained on four datasets with different annotation protocols, the FMB-loss is introduced to facilitate joint training and mitigate the issue of inconsistent gradient magnitudes across datasets. Further details are provided in \textbf{Section III. B}.

\subsection{Fine-grained multi-task balancing loss}
\indent Current facial landmark detection (FLD) methods suffer from the limited scale of available datasets, primarily because annotating facial landmarks is both time-consuming and labor-intensive. To address this problem, we propose a unified training strategy that leverages multiple FLD datasets with diverse annotation protocols. However, this strategy introduces two key challenges:
\textbf{(1) Landmark annotation unification across datasets.} Different datasets often define distinct sets of landmarks with varying numbers and semantic definitions, making it non-trivial to train a unified model. \textbf{(2) Gradient magnitude inconsistency across datasets.} Naively aggregating losses from all datasets can lead to gradient imbalance, where frequently occurring landmarks dominate the optimization process while less frequent ones are underemphasized. This imbalance may result in suboptimal performance on rare landmarks, ultimately degrading overall accuracy. To tackle these challenges, we first unify landmark annotations across different datasets and then propose a Fine-Grained Multi-Task Balance loss (FMB-loss) to alleviate the gradient conflict problem.

\subsubsection{Unified facial landmark definition}
\indent The four widely used FLD datasets are AFLW \cite{koestinger2011annotated}(19 landmarks), WFLW \cite{wu2018look}(98 landmarks), COFW \cite{burgos2013robust}(29 landmarks) and 300W \cite{sagonas2016300}(68 landmarks). Some landmarks are present in all four datasets, while others exist in only one, two, or three datasets. The AFLW dataset's landmark set can be defined as: 
\begin{equation}
A = \left\{a_{1}, a_{2}, \dots,a_{N_{A}}\right\}
\end{equation}
where $N_A=19$ denotes the number of landmarks in AFLW. Similarly, the landmark sets for the other three datasets can be defined accordingly and are denoted by $W$, $C$ and $T$, respectively. Assuming $P$ is a unified facial landmark set, which includes all landmarks appearing in any dataset: 
\begin{equation}
P = \left\{p_{1}, p_{2}, \dots, p_{N_{P}}\right\}
\end{equation}
where $N_P=124$, indicating these four datasets have totally 124 landmarks with different semantic information. There exists a mapping function from each dataset's landmark set $D=\left\{A,W,C,T\right\}$ to the unified landmark set $P$, as well as a mapping function from $P$ back to $D$. These mappings can be formulated as follows:
\begin{align}
& f: D \to P,\quad f\left(X\right)=\left\{p_i\mid p_i\in P,\exists x_j\in X\right\} \\
& g: P \to D,\quad g\left(p_i\right)=\left\{x_j\mid x_j\in X,X\in D\right\}
\end{align}
where $f$ maps landmarks from a specific dataset $X \in D$ to their corresponding landmarks in the unified set $P$, and $g$ retrieves landmarks from $P$ that exist in each dataset. These equations indicate that each landmark in $P$ corresponds to one or more landmarks in $X$, where  $X \in D$.

To represent the occurrence count of a landmark $p_i$ across the four datasets $D=\left\{A,W,C,T\right\}$, we define the counting function $\mathbb{C}\left(p_i\right)$ as follows:
\begin{align}
& \mathbb{C}\left(p_i\right)=\sum_{X \in D}\mathbb{I}\left(p_i\right), p_i \in f\left(X\right) \\
& \mathbb{I}\left(p_i\right)=\left\{
\begin{array}{l}
1, \quad ifp_i \in f(X) \\ 
0, \quad otherwise
\end{array}
\right.
\end{align}
where $\mathbb{I}\left(p_i\right)$ is an indicator function that returns 1 if the landmark $p_i$ in the unified set appears in dataset $X$, and 0 otherwise. The counting function $\mathbb{C}\left(p_i\right)$ determines the number of datasets in which $p_i$ appears.
\begin{figure}[t]
\begin{center}
\includegraphics[width=0.9\linewidth]{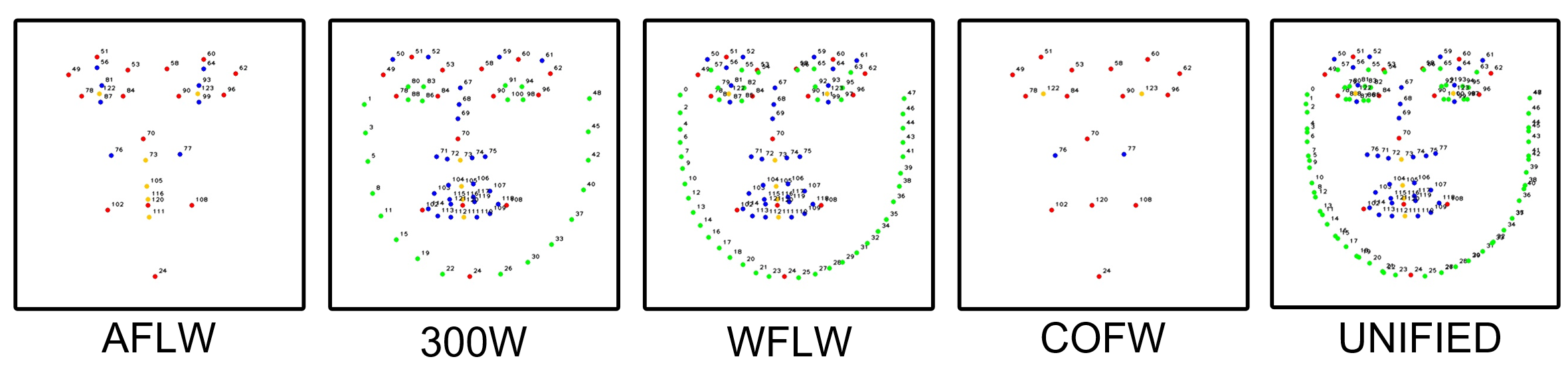}
\end{center}
\vspace{-1em}
	\caption{The landmark annotation distributions across four datasets: AFLW, WFLW, 300W, and COFW. Red dots represent landmarks that appear in all four datasets, while orange, blue, and green denote landmarks shared by three, two, and one dataset(s), respectively. This visualization highlights a challenge in multi-dataset training: landmarks that appear in more datasets tend to receive more attention during learning, while those appearing less frequently are more prone to neglect.}
	\vspace{-1em}
\label{fig3}
\end{figure}
\subsubsection{Effective sample capacity} 
Increasing the amount of training data generally improves model's performance, however, as the data volume grows, information overlap among samples increases, reducing the marginal benefit of additional data \cite{cui2019class}. For the facial landmark detection task, as more data is added, the overlap in annotated landmarks $p_i$ across samples increases. To account for this redundancy, we define the effective sample capacity to measure the amount of unique information during the process of training. The effective sample capacity after the $n$-th sampling $E_n$ is denoted as: 
\begin{equation}
E_n = \frac{1 - \beta^n_{p_i}}{1 - \beta_{p_i}},\text{ where } \beta_{p_i} = \frac{N_{p_i}-1}{N_{p_i}}
\end{equation}
where $N_{p_i}$ represents the effective sample capacity for a given landmark $p_i$ over the entire sampling process. The value of $N_{p_i}$ is primarily related to the diversity among samples. The greater the differences between samples, the larger the value of $N_{p_i}$. The parameter $\beta_{p_i}$ is an intermediate variable introduced for convenience in the mathematical formulation. 

Since the effective sample counts of different landmarks do not vary significantly, we assume that $\beta$ is dataset-dependent and assign the same value of $\beta$ to all landmarks. The $E_n$ can be reformulated as follows:
\begin{equation}
	E_n = \frac{1 - \beta^n}{1 - \beta},\text{ where } \beta = \frac{N-1}{N}
\end{equation}
\indent \textbf{\textit{Proof.}} To facilitate computation, we assume that for a given landmark $p_i$, in the $n$-th sampled data, the landmark either fully belongs to the previously sampled data or does not belong at all, rather than being partially included. This binary assumption provides an intuitive way to characterize the diminishing marginal information phenomenon while simplifying the computation, which in turn facilitates the design of weights and hyperparameter tuning. However, this formulation treats all samples as discrete old/new events, which may somewhat underestimate or overestimate the actual growth of effective sample capacity. In practice, the same landmark under different poses, lighting conditions or expressions often contains both redundant and novel information, and the proportion of these two types of information is difficult to quantify precisely. Therefore, we adopt this idealized assumption as an approximation to obtain a tractable theoretical framework. Despite this simplification, the derived effective sample capacity formula still reasonably captures the trend of diminishing marginal information gain as more samples are added, which forms the theoretical basis for applying landmark-level inverse-frequency weighting in the FMB-loss.
\\\indent Initially, after the first sampling, the sample is completely novel and does not overlap with any previous samples, so $E_1=1$, assume that after the $(n-1)$-th sampling, the effective sample capacity is:
\begin{equation}
E_{n-1}=\frac{1-\beta^{n-1}}{1-\beta}
\end{equation}
\indent During the $n$-th sampling, the probability that the sampled landmark already exists in previous samples is $p=E_{n-1}/N$ and the probability of sampling a new landmark is $1-p$. Therefore, after the $n$-th sampling, the updated effective sample capacity $E_n$ is:
\begin{equation}
\begin{aligned}
E_n&=pE_{n-1}+\left(1-p\right)\left(E_{n-1}+1\right)
\\ &=1-p+E_{n-1}
\\ &=1+\frac{N-1}{N}\frac{1-\beta^{n-1}}{1-\beta}
\\ &=\frac{1-\beta^n}{1-\beta}
\end{aligned}
\end{equation}
\indent This result captures the diminishing gain in unique information as more samples are added, governed by the redundancy ratio $\beta$.
\subsubsection{FMB-loss}
\indent In multi-task joint training, the model may pay varying degrees of attention to different tasks, often focusing excessively on certain tasks while neglecting others. This imbalance can result in the model performing well on specific tasks but poorly on the overall task. Moreover, inconsistency in task weighting may result in uneven convergence rates, where tasks assigned greater weights converge more rapidly, while others lag behind, thereby hindering overall training efficiency. For multi-task joint training,  the overall loss function can generally be formulated as follows:

\begin{equation}
	\mathcal{L}=\sum_{X \in D} \omega_{X}L_{X}
\end{equation}
where $\omega_{X}$ is the weight of the $X$-th task/dataset and $L_X$ denotes the corresponding loss. Current methods\cite{kendall2018multi}, \cite{liu2019end} introduce weights  $\omega_{X}$ to balance multiple tasks, but in facial landmark detection across different datasets, there exist direct correlations among the datasets, as they share common landmarks, as shown in Fig. \ref{fig3}. A coarse-grained task-level balancing approach would overlook these inter-task relationships. To fully exploit these correlations and achieve a finer-grained balance, we introduce a landmark-level balancing strategy for unified training across multiple datasets. Specifically, we propose an alternative approach that adaptively adjusts weights based on the frequency of landmark occurrences during training. First, the AWing Loss \cite{wang2019adaptive} is used to compute the heatmap-based loss for each landmark $x_j$ in dataset $X$:
\begin{equation}
	\mathcal{L}_{awing}^{x_j} = 
	\begin{cases}
		\omega \ln\left(1 + \left|\frac{y - \hat{y}}{\epsilon}\right|^{\alpha - y} \right), & \text{if } |y - \hat{y}| < \theta \\
		A |y - \hat{y}| - C, & \text{otherwise}
	\end{cases}
\end{equation}
where $y$ and $\hat{y}$ denote the predicted landmark heatmap and ground truth one, respectively. $A=\omega(1/(1+(\theta/\epsilon)^{(\alpha-y)}))(\alpha-y)((\theta/\epsilon)^{(\alpha-y-1)})(1/\epsilon)$, $C=(\theta A-\omega ln(1+(\theta/\epsilon)^{\alpha-y}))$. And $\omega=14$, $\theta=0.5$, $\alpha=2.1$, $\epsilon=1$. Next, we introduce the inverse of the effective sample capacity to balance the loss. Specifically, for each landmark $p_i$ in the unified facial landmark set $P$, the effective sample capacity is defined as:
\begin{equation}
	E_{p_i}=\frac{1-\beta^{\mathbb{C}\left(p_i\right)}}{1-\beta}
\end{equation}
where $\beta$ is a tunable parameter with values in $[0,1)$. When $\beta = 0$, $E_{p_i}$ is 1, and when beta approaches 1, $E_{p_i} \to \mathbb{C}\left(p_i\right)$. The limiting case is as follows:
\begin{equation}
	\lim\limits_{\beta \to 1}E_{p_i}=\lim\limits_{\beta \to 1}\frac{1-\beta^{\mathbb{C}\left(p_i\right)}}{1-\beta}
	=\lim\limits_{\beta \to 1}\frac{\left(1-\beta^{\mathbb{C}\left(p_i\right)}\right)^\prime}{\left(1-\beta\right)^\prime}=\mathbb{C}\left(p_i\right)
\end{equation}
\indent The loss for each landmark is then weighted by the inverse of $E_{p_i}$, resulting in the balanced landmark loss:
\begin{equation}
	\mathcal{L}_{bal}^{x_j}=\frac{1}{E_{p_i}}\mathcal{L}_{awing}^{x_j}, \exists f\left(x_j\right)=p_i
\end{equation}
where $\mathcal{L}_{bal}^{x_j}$ denotes the loss of the $j$-th landmark in dataset $X$ and $f\left(x_j\right)$ maps the sample $x_j$ to its corresponding landmark $p_i$. This indicates that a smaller value of $\beta$ corresponds to little or no reweighting, while a larger value enhances the balancing effect by approximating the inverse of the number of times a landmark is used during an iteration. Finally, the overall balanced loss function is formulated as:
\begin{equation}
	\mathcal{L}_{bal}=\frac{1}{n} \sum_{X \in D}\frac{\sum_{j=1}^{N_X}\mathcal{L}_{bal}^{x_j}}{N_X}
\end{equation}
where $n$ denotes the number of samples in a mini-batch.

The proposed FMB-loss introduces a landmark-level balancing strategy by leveraging the concept of effective sample capacity. Unlike conventional approaches that treat each dataset or task equally, FMB-loss adaptively adjusts the contribution of each landmark based on its frequency of occurrence. A key component of this strategy is a tunable parameter $\beta$, which controls the sensitivity of the weighting scheme, thereby alleviating the gradient conflict and improving the landmark detection accuracy.

\subsection{Frequency-Guided Structure-Aware(FGSA) model}

Heatmap regression-based methods \cite{ma2022robust}, \cite{yang2017stacked}, \cite{dapogny2019decafa} have achieved great success in facial landmark detection (FLD) due to their strong spatial generalization ability. However, under challenging scenarios such as large pose variations or complex illumination, their capacity to capture facial structural information is substantially diminished, leading to the loss of critical structural cues and consequently inaccurate landmark detection. To address this issue, we propose enhancing the model's perception of facial structure by explicitly incorporating high-frequency information, which captures rapid intensity variations and typically corresponds to structural details such as boundaries where most landmarks reside. By emphasizing these features, the model can obtain more robust structural priors, enabling accurate landmark detection even in the presence of extreme poses and expressions. Building on these insights, we introduce a novel Frequency-Guided Structure-Aware (FGSA) model that incorporates high-frequency information to improve facial landmark detection. By enhancing the representation of facial boundary and shape-related cues, the FGSA model improves the model's ability to detect landmarks under adverse conditions.
\begin{figure*}[!t]
	\centering
	\includegraphics[width=0.9\linewidth]{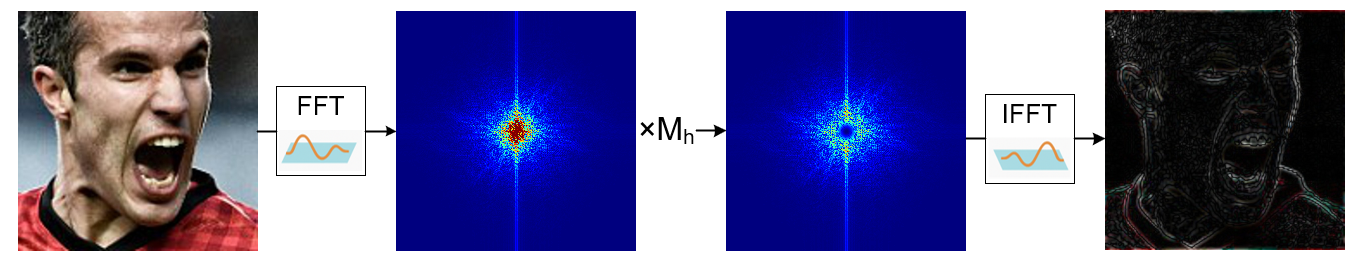}
	\centering
	\caption{The high-frequency information extraction. Given an input image, it is first transformed into the frequency domain via the Fast Fourier Transform (FFT). A high-frequency emphasis mask $M_h$ is then applied element-wise to the spectrum, followed by an inverse FFT to reconstruct a spatial-domain image $I_{hf}$ that predominantly contains high-frequency components. For example, the resulting image clearly highlights the structural information of faces with large poses, thereby providing valuable cues for enhancing FLD performance.}
	\label{fig1}
	\vspace{-1em}
\end{figure*}
\\\indent High-frequency information in an image corresponds to regions with rapid variations across different directions, typically representing edges and details. These information are crucial for facial landmark detection, as most landmarks lie along facial contours and boundaries. Assuming $FFT$ denotes the Fast Fourier Transform and $IFFT$ denotes its inverse. Given an image $I$, its frequency-domain representation after applying $FFT$, and then the original image can be reconstructed from its frequency representation using $IFFT$:
\begin{equation}
	f = FFT\left(I\right)
\end{equation}
\begin{equation}
	I = IFFT\left(f\right)
\end{equation}

To extract high-frequency components from the image, we apply a high-frequency emphasis mask to the image spectrum, defined as follows:
\begin{equation}
	M_h^{i,j} = 1-exp\left(-\frac{\left(i-\frac{W}{2}\right)^2+\left(j-\frac{H}{2}\right)^2}{2\sigma^2}\right)
\end{equation}
where $\sigma$ is a hyperparameter that defines the mask region. $H$ and $W$ represent the height and width of the image, respectively. This mask follows a Gaussian distribution, where values near the image center are close to zero and gradually increase toward the edges, reaching values close to 1. By adjusting $\sigma$, the region of emphasis can be effectively controlled. With the mask $M_h$, the enhanced high-frequency information $I_{hf}$ can be extracted as follows:
\begin{equation}
	I_{hf} = IFFT\left(fM_h\right)
\end{equation}
\indent As illustrated in Fig. \ref{fig1}, with the extracted high-frequency information, a high-frequency image can retain critical facial boundaries and structure, even under large pose variations. These structure-enhanced representations are especially useful for facial landmark detection, as landmarks are often located along facial contours, including the eyelids, lips and jawline. The reconstructed high-frequency image is further injected and regularized for enhancing facial structure. 


\textbf{Frequency-guided structure injection.} Accurate facial landmark detection requires effective modeling of structural cues across multiple spatial scales. To this end, we propose a frequency-guided structure injection module that captures and injects multi-scale facial structure information into the Transformer network. It begins by processing the high-frequency information $I_{hf}$ to obtain multi-scale high-frequency features. Let $F^{i}_{s}$ represent the stage-specific structure features corresponding to the stage $i$, where $i \in \{1, 2, 3, 4\}$. This process can be defined as:
\begin{align}
	F^{i}_{s} &= Refine^{i}\left(F^{i-1}_{s}\right)
\end{align}
where $Refine$ consists of several convolutional layers, followed by batch normalization and ReLU activation functions, and $F^{0}_{b}=I_{hf}$. On the other hand, the structural information is closely related to the original image. Incorporating original image features further enhances the model's ability to capture semantic image context. Therefore, we also refine the original features. $F^{i}_{pa}$ denotes the image information appropriate for each stage, where $i \in \{1, 2, 3, 4\}$, and the refining process is formulated as follows:
\begin{align}
	F^{i}_{pa} &= Linear^{i}\left(F^{i-1}_{pa}\right), 
\end{align}
where $F^{0}_{pa}=I_{pa}$. Both the $Refine$ and $Linear$ operations serve the purpose of aligning the features to match the spatial dimensions at different stages. Then, the structure features $F^{i}_{s}$ and the refined image features $F^{i}_{pa}$ are concatenated to generate the facial structure prompts:
\begin{equation}
	F^{i}_{p} = Concat\left(F^{i}_{pa}, F^{i}_{s}\right)
\end{equation}

By combining $F^{i}_{s}$ and $F^{i}_{pa}$, the proposed FGSA model can benefit from both high-frequency structural cues and global semantic context. These facial structure prompts are injected into each stage of the original Transformer network:
\begin{equation}
	F^{i,j}_{in} = Concat\left(F^{i}_{p}, F^{i,j}\right)
\end{equation}
where $F^{i,j}$ denotes the input feature of the $j$-th Transformer layer in the $i$-th stage.

\textbf{Frequency-guided structure regularization.} While the injection of facial structure prompts strengthens the model's ability to perceive facial structure, it may simultaneously introduce redundant or noisy information,  which could hinder stable optimization and lead to overfitting. To address this problem, we introduce a structure regularization that not only refines the injected features but also improves the model's generalization and robustness. The structure regularization is processed by a shared channel attention module \cite{woo2018cbam} within each stage, which helps select the most informative channels. Next, it introduces two types of MLPs: A shared MLP (denoted as $MLP^{i,j}_{us}$) across all layers in the same stage, which captures common information across transformer layers. The index $i$ refers to the $i$-th stage, while $j$ refers to the $j$-th Transformer layer corresponding to that stage. This helps stabilize and simplify training while acting as a form of regularization to prevent overfitting. An independent MLP (denoted as $MLP^{i}_{s}$ ) for each transformer layer, which enables fine-grained adjustments at different layers, enhancing the model's expressiveness and feature fusion capabilities. The whole structure regularization process can be formulated as:
\begin{equation}
	F^{i,j}_{in} = MLP^{i}_{s}\left(GELU\left(MLP^{i,j}_{us}\left(CA^i\left(F^{i,j}_{in}\right)\right)\right)\right)
\end{equation}

This frequency-guided structure regularization facilitates the selective integration of boundary prompts, enabling them to effectively contribute to robust and accurate facial structure modeling.

\section{Experiments}
\indent In this section, we conduct detailed experiments on four datasets, and the details are as follows.
\subsection{Dataset and implementation details}
\indent \textbf{300W}(68 landmarks)\cite{sagonas2016300}: It consists of 3,148 images for training and 689 images for testing. The training set is composed of images from LFPW, AFW and HELEN. The test set includes images from IBUG, along with the test subsets of HELEN and LFPW. The test set is further divided into two subsets: (1) the common subset, which contains 554 images (330 from the HELEN test set and 224 from the LFPW test set); (2) the challenging subset, which consists of 135 images from the IBUG dataset, featuring more complex real-world scenarios.
\\\indent \textbf{WFLW}(98 landmarks)\cite{wu2018look}: WFLW is a facial landmark dataset containing 10,000 images, with 7,500 images used for training and 2,500 images for testing, each annotated with 98 facial landmarks. The dataset encompasses a diverse range of real-world variations, including different poses, expressions, lighting conditions, occlusions, and blurriness. In addition to landmark annotations, WFLW provides rich attribute annotations, such as occlusion, head pose, blur, makeup, illumination and expression, enabling comprehensive evaluation of facial landmark detection algorithms under challenging conditions.
\\\indent \textbf{COFW}(29 landmarks)\cite{burgos2013robust}: COFW is a challenging facial landmark dataset specifically designed to address occlusion issues. It contains 1,345 training images, including 845 images from the LFPW training set and an additional 500 heavily occluded faces. The test set consists of 507 images with significant occlusions, large pose variations, and diverse expressions. Each image is annotated with 29 facial landmarks, providing a benchmark for evaluating landmark detection algorithms under challenging conditions.
\\\indent \textbf{AFLW}(19 landmarks)\cite{koestinger2011annotated}: AFLW is a large-scale facial landmark dataset containing 25,993 images with diverse variations in pose, occlusion, and background. Each face is annotated with 19 landmarks, making it a valuable resource for evaluating face alignment methods under challenging conditions. The dataset includes extreme head poses, with yaw angles ranging from -120$^\circ$ to +120$^\circ$ and pitch angles from -90$^\circ$ to +90$^\circ$. AFLW is typically divided into two subsets: AFLW-Full, which consists of 24,386 images split into 20,000 for training and 4,386 for testing, and AFLW-Frontal, which selects a subset of 1,165 frontal-view images from the test set to specifically assess alignment performance on front-facing faces.
\begin{table}
	\caption{Comparisons with state-of-the-art methods on the 300W dataset. The error ($NME_{io}$) is normalized by the inter-ocular distance. (\% omitted)}
	\begin{center}
		\begin{tabular}{p{4.3cm}p{2cm}p{2cm}p{1cm}}
			\hline
			Method & Common  & Challenging  & Full  \\
			\hline
			HGs(ECCV16)\cite{yang2017stacked}&	3.72&	7.23&	4.41\\	
			MDM(CVPR16) \cite{trigeorgis2016mnemonic}&	4.36&	7.56&	4.99\\
			LAB(CVPR18) \cite{wu2018look}&	2.98&	5.19&	3.49\\
			ODN(CVPR19)\cite{zhu2019robust} &	3.56&	6.67&	4.17\\
			AWing(ICCV19)\cite{wang2019adaptive}&	2.72&	4.52&	3.07\\
			LUVLi(CVPR20)\cite{kumar2020luvli} &	2.76	&5.16&	3.23\\
			ADNet(ICCV21)\cite{huang2021adnet} &	2.53&	4.58&	2.93\\
			DTLD(CVPR22)\cite{li2022towards} &2.59 & 4.50& 2.96\\
			SLPT(CVPR22)\cite{xia2022sparse}&	2.75&	4.90&	3.17\\
			GlomFace(CVPR22)\cite{zhu2022occlusion} &	2.72&	4.79&	3.13\\
			PicassoNet(TNNLS23)\cite{wen2022picassonet}&	3.03& 5.81&  3.58\\
			GFL(CVPR24)\cite{liang2024generalizable} &	2.79&	4.91&	3.20\\
			\hline
			\textbf{FGTBT(ours)}&	\textbf{2.71}&	\textbf{4.51}&	\textbf{3.06}\\
			\hline
		\end{tabular}
	\end{center}
	\label{tab300w}
\end{table}
\\\indent \textbf{Evaluation metrics.} Normalized Mean Error (NME) is a widely adopted evaluation metric for FLD. NME is calculated as follows:
\begin{equation}
NME\left(Y, \hat{Y}\right)=\frac{1}{N_p}\sum_{i=1}^{N_p} \frac{\| y_i-\hat{y_i}\|_2}{d}
\end{equation}
where $N_p$ denotes the number of landmarks in a given dataset, $y_i$ and $\hat{y_i}$ represent the predicted coordinates and ground-truth of the $i$-th landmark, respectively. $d$ is the normalization factor. For the 300W and WFLW datasets, the inter-ocular distance is used as the normalization term; for the AFLW dataset, the face size is used; and for the COFW dataset, the distance between the outer corners of the eyes is adopted. 
\\\indent We also evaluate model performance using the Failure Rate (FR), which measures the proportion of samples whose NME exceeds a predefined threshold $\tau$, which could be defined as follows:
\begin{equation}
FR\left(Y, \hat{Y}\right)=\frac{1}{N_p}\sum_{i=1}^{N_p}1_{\left\{NME\left(Y_i, \hat{Y_i}\right)>\tau\right\}}
\end{equation}
where $N_p$ denotes the number of landmarks in a given dataset, $1_{\left\{.\right\}}$ denotes the indicator function, which takes the value 1 if the specified condition is satisfied and 0 otherwise. In our experiments, the threshold $\tau$ is set to 0.10. 
\begin{table}
	\caption{Comparisons with state-of-the-art methods on WFLW dataset. The error ($NME_{io}$) is normalized by the inter-ocular distance. (\% omitted)}
	\begin{center}
		\resizebox{15.5cm}{!}{
		\begin{tabular}{p{4.8cm}p{1.1cm}p{1.2cm}p{1.5cm}p{1.7cm}p{1.6cm}p{1.4cm}p{1.3cm}}
			\hline
			Method & Testset  & Pose Subset  & Expression Subset &Illumination Subset &Make-Up Subset &Occlusion Subset & Blur Subset  \\
			\hline
			\makecell[l]{LAB(CVPR18)\cite{wu2018look}} &5.27 &10.24 &5.51 &5.23 &5.15 &6.79 &6.32 \\
			\makecell[l]{DeCaFA(ICCV19)\cite{dapogny2019decafa}}&	4.62&	8.11&	4.65&	4.41&	4.63&	5.74&	5.38\\
			\makecell[l]{HRNet(TPAMI20)\cite{wang2020deep}}&4.60 &7.86 &4.78 &4.57 &4.26 &5.42 &5.36 \\
			\makecell[l]{LUVLi(CVPR20)\cite{kumar2020luvli}}&4.37 &7.56 &4.77 &4.30 &4.33 &5.29 &4.94 \\
			\makecell[l]{MMDN(TNNLS21)\cite{wan2021robust}}& 4.87& 7.71& 4.79& 4.61& 4.72& 6.17& 5.72\\
			\makecell[l]{GlomFace(CVPR22)\cite{zhu2022occlusion}}& 4.81& 8.71& -& -& -& 5.14& -\\ 
			\makecell[l]{EfficientFAN(TNNLS23)\cite{gao2021facial}}& 4.54& 8.2& 4.87& 4.39& 4.54& 5.42& 5.04\\ 
			\makecell[l]{PicassoNet(TNNLS23)\cite{wen2022picassonet}}& 4.82& 8.61& 5.14& 4.73& 4.68& 5.91& 5.56\\ 
			\hline
			\makecell[l]{\textbf{FGTBT}  \textbf{(ours)}}&	\textbf{4.42}&	\textbf{7.43}&	\textbf{4.54}& \textbf{4.47}&	\textbf{4.37}&	\textbf{5.48} & \textbf{5.02}\\
			\hline
		\end{tabular}}
	\end{center}
	\label{tabwflw}
\end{table}
\\\indent \textbf{Implementation details.} In our experiments, all images were first cropped to a resolution of $480 \times 480$. To augment the training data, we applied the following strategies: (1) random scaling within the range of 1.0$\times$ to 1.25$\times$, (2) randomly selecting 60\% of the images for in-plane rotation by a random angle between -30$^{\circ}$ and +30$^{\circ}$ counterclockwise, and (3) horizontally flipping 50\% of the images. The FGTBT model was trained for a total of 100,000 iterations. The initial learning rate was set to 2.5e-4 and was decayed to 80\% of its current value at the 40,000th, 70,000th, and 90,000th iterations. In each training iteration, two samples were randomly drawn from each of the four datasets and concatenated along the batch dimension to form a mixed batch input. The loss was computed independently for each dataset based on its corresponding output. The FGTBT model was trained on a single NVIDIA GeForce RTX 4090 GPU. For the hyperparameter settings, the values of $\omega$, $\theta$, $\alpha$, and $\epsilon$ in $\mathcal{L}_{awing}^{X_i}$ are set to 14, 0.5, 2.1, and 1, respectively. For the $\sigma$ in FGSA model is set to 20. The $\beta$ in FMB-loss is set to 0.9.

\subsection{Evaluation under normal circumstances}
Under normal circumstances, where faces are free from significant occlusions, pose variations, expressions, illumination or blur. We evaluate the corresponding performance of FGTBT on three datasets: AFLW fullset, 300W-common and 300W-fullset. As shown in Table \ref{tab300w}, FGTBT achieves an $NME_{io}$ of 2.71\% on the 300W common subset, outperforming most of current methods \cite{trigeorgis2016mnemonic}, \cite{kumar2020luvli}, \cite{xia2022sparse}, \cite{liang2024generalizable}. On the 300W-fullset, it achieves an $NME_{io}$ of 3.06\%, again surpassing most existing benchmarks \cite{zhu2019robust}, \cite{yang2017stacked}, \cite{wu2018look}, \cite{wang2019adaptive}, \cite{zhu2021improving}, \cite{zhu2022occlusion}. According to Table \ref{tabaflw}, FGTBT also yields better performance on the AFLW dataset compared to \cite{feng2017dynamic}, \cite{wu2018look}. It is worth noting that the proposed FGTBT performs slightly worse than \cite{jin2021pixel}, \cite{kumar2020luvli} on the AFLW dataset, mainly because these methods are trained on only 19 landmarks, which enables them to focus on a smaller set of landmarks and thereby achieve better performance.

\subsection{Evaluation of robustness against large poses and expressions}
Faces with large poses or exaggerated expressions exhibit complex and challenging geometric constraints, which are difficult to model and consequently lead to performance degradation. To evaluate FGTBT under such conditions, we conducted experiments on WFLW, AFLW fullset and the 300W- chall subset.

On the 300W-chall subset, FGTBT achieves an $NME_{io}$ of 4.51\%, as shown in Table \ref{tab300w}, outperforming current methods \cite{zhu2021improving}, \cite{zhu2022occlusion}, \cite{xia2022sparse}, \cite{liang2024generalizable}. It also shows lower error rates on the AFLW fullset as shown in Table \ref{tabaflw} and excels in the pose and expression subsets of WFLW as shown in Table \ref{tabwflw} compared to \cite{wu2018look}, \cite{dapogny2019decafa}, \cite{wang2020deep}, \cite{wan2021robust}, \cite{zhu2022occlusion}, further validating its robustness in extreme facial variations. The above experimental result indicates that: 1) The proposed FGSA model explicitly captures and models facial structure cues, enhancing robustness to faces under facial expression variations or large pose variations, and 2) the FMB-loss harmonizes datasets with different annotation protocols, enabling the model to leverage the facial semantic landmark annotations from various datasets during training, thereby facilitating more effective facial structure modeling.
\begin{table}
	\caption{Comparisons with state-of-the-art methods on the COFW dataset. The error ($NME_{ip}$) is normalized by the inter-pupil distance. (\% omitted)}
	\vspace{-1em}
	\begin{center}
		\begin{tabular}{p{6cm}p{1.6cm}p{2.2cm}}
			\hline
			Method & $NME_{ip}$  & FR \\
			\hline
			AWing(ICCV19) \cite{wang2019adaptive}&	4.94&	0.99\\
			ODN(CVPR19) \cite{zhu2019robust}&		5.30&	-\\
			ADNet(ICCV21) \cite{huang2021adnet}&	4.68&	0.59\\
			MMDN(TNNLS22)  \cite{wan2021robust}&	5.01&	1.78\\
			SLPT(CVPR22)  \cite{xia2022sparse}&	4.79&	1.18\\
			DSLPT-R50(TPAMI23)  \cite{xia2023robust}&	4.81&	1.18\\
			CIT-v2(IJCV24)  \cite{li2024cascaded}&	5.81&	3.55\\
			\hline
			\textbf{FGTBT(ours)}&	\textbf{4.79} &\textbf{0.59}\\
			\hline
		\end{tabular}
	\end{center}
	\vspace{-2em}
	\label{tabcofw}
\end{table}
\begin{table}
	\caption{Comparisons with state-of-the-art methods on the AFLW dataset. The error ($NME_{diag}$) is normalized by face size. (\% omitted)}
	\begin{center}
		\begin{tabular}{p{5cm}p{2.5cm}}
			\hline
			Method & Full Testset \\
			\hline
			DAC-CSR(CVPR17) \cite{feng2017dynamic}&	2.27 \\
			LAB(CVPR18) \cite{wu2018look}&	1.85\\
			LUVLi(CVPR20) \cite{kumar2020luvli}& 1.39 \\
			HRNet(TPAMI20) \cite{wang2020deep}& 1.57\\
			PIPNET(IJCV21) \cite{jin2021pixel}&	1.42\\
			PicassoNet(TNNLS23) \cite{wen2022picassonet} &1.59\\
			\hline
			\textbf{FGTBT(ours)}&	\textbf{1.67}	\\
			\hline
		\end{tabular}
	\end{center}
	\label{tabaflw}
\end{table}

\subsection{Evaluation of robustness against illumination and blur}
Faces captured under low-light or blurred conditions present significant challenges due to the loss of visual detail. To evaluate the robustness of our approach in these adverse scenarios, we perform assessments on the 300W-chall subset, along with the Illumination and Blur subsets of the WFLW dataset.
\\\indent On the 300W-chall subset, as shown in Table \ref{tab300w}, FGTBT achieves an $NME_{io}$ of 4.51\%, outperforming competing methods. In the illumination and blur subsets of WFLW, as shown in Table \ref{tabwflw}, FGTBT attains $NME_{io}$ of 4.47\% and 5.02\%, respectively, surpassing methods such as \cite{wu2018look}, \cite{dapogny2019decafa}, \cite{wang2020deep}, \cite{wan2021robust}. These results suggest that FGTBT maintains high accuracy even under poor image quality. This robustness is largely attributed to the proposed FGSA model, which can effectively preserve facial structural cues even when the facial image is degraded by blur and illumination.

\subsection{Evaluation of robustness against occlusion}
\indent Occlusion significantly affects FLD performance, as most models struggle to capture complete semantic information when parts of the face are blocked. To evaluate the robustness of our proposed FGTBT under occlusion scenarios, we conduct experiments on the 300W-chall subset, the occlusion subset of the WFLW dataset, and the COFW dataset.
\\\indent As shown in Table \ref{tabcofw}, on the COFW dataset, FGTBT achieves an $NME_{ip}$ of 4.79\% and a FR of 0.59\%, outperforming current methods \cite{wang2019adaptive}, \cite{zhu2019robust}, \cite{wan2021robust}, \cite{xia2022sparse}, \cite{li2024cascaded}.  The performance is slightly inferior to ADNet\cite{huang2021adnet}, primarily because ADNet is trained with only 29 landmarks, enabling it to focus on a smaller prediction space, which naturally leads to higher accuracy. As shown in Table \ref{tabwflw}, on the WFLW occlusion subset, the model attains an $NME_{io}$ of 5.48\%, showing a improvement over existing approaches \cite{wu2018look}, \cite{dapogny2019decafa}, \cite{wan2021robust}. Similarly, in the 300W-chall dataset, FGTBT demonstrates high stability and accuracy compared to other methods \cite{zhu2021improving}, \cite{zhu2022occlusion}, \cite{xia2022sparse}, \cite{liang2024generalizable}. These results confirm that FGTBT is highly robust to occlusion. This robustness can be mainly attributed to our proposed FMB-loss, which enables the joint training of four differently annotated datasets within a unified framework. As a result, FGTBT can capture richer and more comprehensive facial structural information, which significantly enhances its ability to detect landmarks even under severe occlusions.

\subsection{Self evaluation}
\begin{figure*}[!t]
\begin{center}
\includegraphics[width=\linewidth]{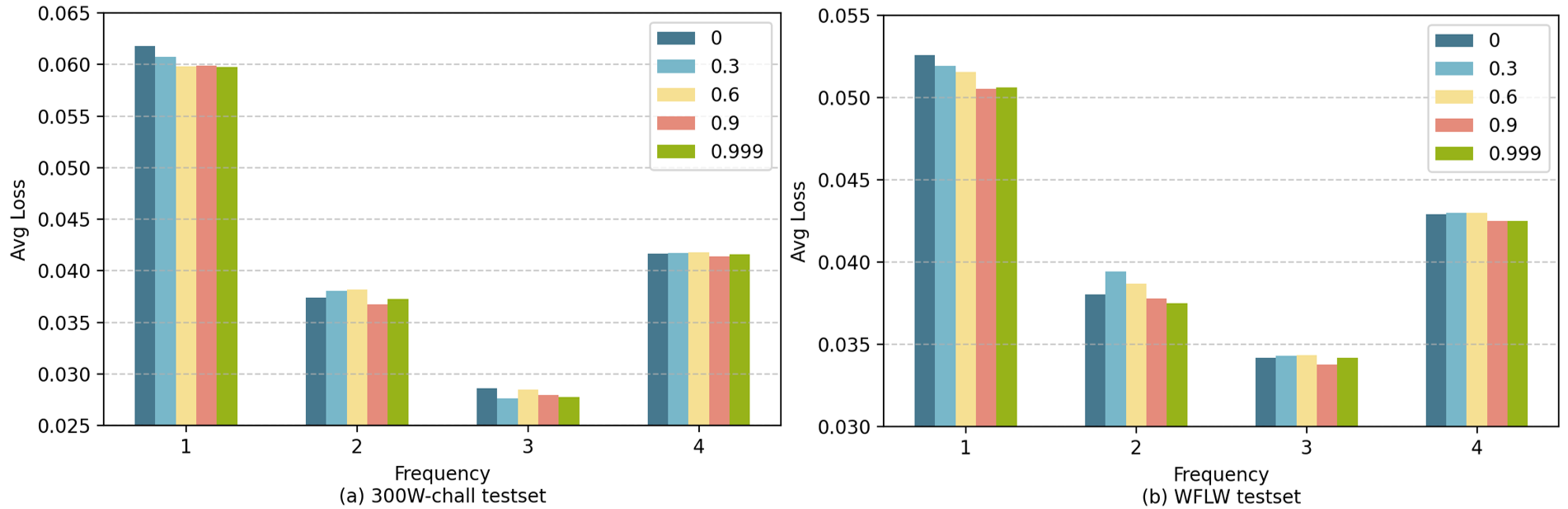}
\end{center}
\vspace{-1em}
	\caption{The average loss of landmarks that appear with different frequencies across the four datasets under different values of $\beta$. (a) presents the results on the 300W-chall testset, while (b) shows the results on the WFLW testset.}
\label{fig4}
\end{figure*}

\textbf{Evaluation of the hyperparameter $\beta$ in FMB-loss.} We study the effect of $\beta$ in FMB-loss. When $\beta=0$, no reweighting is applied; as $\beta$ increases, more emphasis is placed on infrequent landmarks. Small $\beta$ values improve accuracy on frequent landmarks but hurt rare ones, while large $\beta$ values have the opposite effect. As shown in Table \ref{beta} and Fig. \ref{fig4}, increasing $\beta$ from 0 to 0.9 yields significant improvements on 300W-chall (2.1\%) and WFLW (2.4\%), while performance on COFW and AFLW remains stable since they already contain mainly low-frequency landmarks. However, excessively large $\beta$ (e.g., 0.999) slightly degrades performance, indicating that over-emphasizing rare landmarks weakens the model’s ability to capture common but semantically important ones. Overall, a moderate $\beta$ provides the best trade-off and improves generalization across datasets.

It is worth noting that the loss distribution varies significantly across landmarks with different occurrence frequencies, influenced by dataset definitions and the model's ability to localize specific landmarks. Landmarks with clear semantic cues are easier to localize and yield lower losses, while those lacking such cues are more difficult and lead to higher losses. As shown in Fig. \ref{fig3}, landmarks 0-23 and 25-47 generally lack explicit semantic meaning, resulting in higher losses, whereas landmarks appearing three times are all semantically clear and thus achieve the lowest losses. In contrast, landmarks appearing four times show mixed cases, with some (e.g., 51 and 60) being harder to localize. Overall, the variation in loss is largely determined by the intrinsic difficulty of the landmarks.

\begin{table}
	\caption{THE EFFECT OF $\beta$ IN FMB-LOSS ON BENCHMARK DATASETS (NME(\%))}
	\vspace{-1em}
	\begin{center}
		\begin{tabular}{p{3.5cm}p{2cm}p{2cm}p{1.2cm}p{1.2cm}}
			\hline
			\diagbox{$\beta$}{Testset} & 300W-chall & WFLW-full & COFW & AFLW  \\
			\hline
			0& 4.61& 4.53& 4.84& \textbf{1.66}\\
			0.3& 4.62& 4.54& \textbf{4.76}& \textbf{1.66}\\
			0.6& 4.55& 4.47& 4.83& \textbf{1.66}\\
			0.9& \textbf{4.51}& \textbf{4.42}& 4.79& 1.67\\
			0.999& 4.53& \textbf{4.42}& 4.83& 1.67\\
			\hline
		\end{tabular}
	\end{center}
	\vspace{-1em}
	\label{beta}
\end{table}
\begin{figure*}[!t]
\begin{center}
\includegraphics[width=0.7\linewidth]{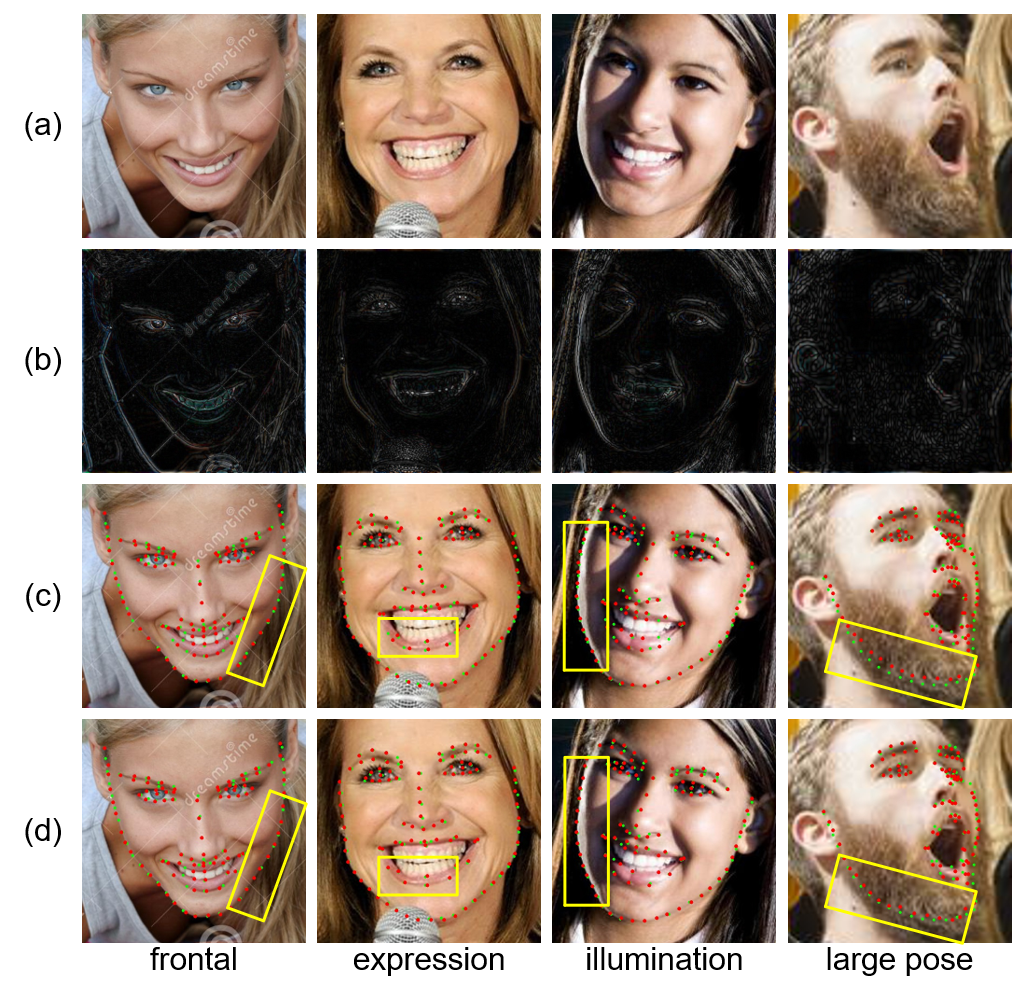}
\end{center}
\vspace{-1em}
	\caption{(a) The original input images, (b) the high-frequency image, (c) the prediction results of the FGTBT without the FGSA model, and (d) the prediction results with the FGSA model integrated. It can be seen that by integrating FGSA model, the FGTBT can learn more effective facial structural information, thus improving the landmark detection accuracy.}
	\vspace{-1em}
\label{fig5}
\end{figure*}
\begin{figure*}[!t]
\begin{center}
\includegraphics[width=0.9\linewidth]{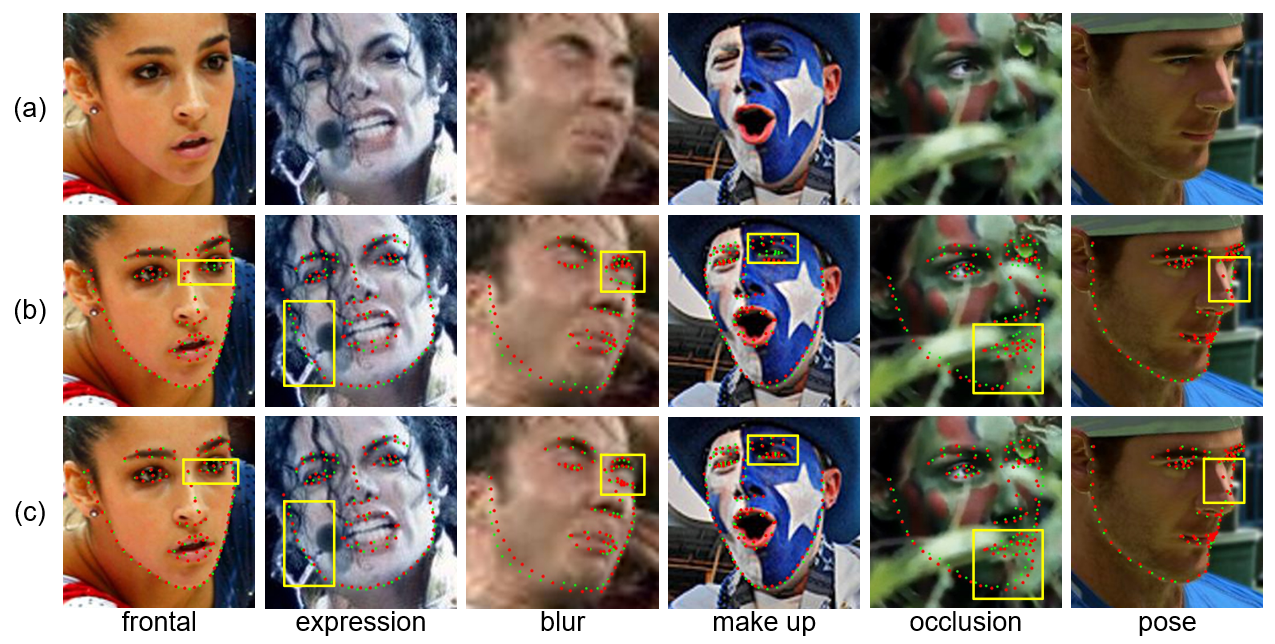}
\end{center}
\vspace{-1em}
	\caption{(a) The original input images, (b) the prediction results of the FGTBT without the FMB-loss and (c) the prediction results with the FMB-loss integrated. It can be seen that by integrating FMB-loss, the model can more effectively leverage information from multiple datasets, thereby detecting more accurate landmarks.}
	\vspace{-1em}
\label{fig7}
\end{figure*}
\begin{table}
	\caption{COMPARISONS WITH OTHER MTL BALANCE METHOD ON BENCHMARK DATASETS (NME(\%))}
	\vspace{-1em}
	\begin{center}
		\begin{tabular}{p{3.5cm}p{2cm}p{2cm}p{1.2cm}p{1.2cm}}
			\hline
			\diagbox{Method}{Testset} & 300W-chall & WFLW-full & COFW & AFLW  \\
			\hline
			baseline&	4.63& 4.64& 4.84& 1.67\\
			DWA\cite{liu2019end}& 4.61& 4.72& 4.92& 1.69\\
			UW\cite{kendall2018multi}& 4.58& 4.71& 4.94& 1.68\\
			\hline
			\textbf{FMB(ours)}& \textbf{4.51}& \textbf{4.47}& \textbf{4.83}& \textbf{1.68}\\
			\hline
		\end{tabular}
	\end{center}
	\vspace{-1em}
	\label{mtl}
\end{table}
\begin{table}
	\caption{COMPARISONS OF DIFFERENT SCHEMES FOR INTRODUCING FREQUENCY-DOMAIN FEATURES}
	\vspace{-1em}
	\begin{center}
		\begin{tabular}{p{3.5cm}p{2cm}p{2cm}p{1.2cm}p{1.2cm}}
			\hline
			\diagbox{Method}{Testset} & 300W-chall & WFLW-full & COFW & AFLW  \\
			\hline
			SRM\cite{han2021fighting}&4.49&4.43 &4.87 &1.68 \\
			Wavelet\cite{yuan2025feature}& 4.59& 4.48& 4.84&1.66 \\
			\hline
			\textbf{FFT}\cite{heckbert1995fourier}\textbf{(ours)}& \textbf{4.51}& \textbf{4.42}& \textbf{4.79}& \textbf{1.68}\\
			\hline
		\end{tabular}
	\end{center}
	\vspace{-1em}
	\label{frequency}
\end{table} 
\textbf{Analysis of the FGSA-model and FMB-loss.} For faces with expressions, large pose variations or illumination, the FGSA model can help model facial structure by introducing high-frequency information, which in turn improves the landmark detection accuracy. We visualize the corresponding experimental result in Fig. \ref{fig5}. Note that to make the images clearer, the extracted $I_{hf}$ is normalized as:
\begin{equation}
I_{i,j}^{new}=\frac{I_{i,j}-I_{min}}{I_{max}-I_{min}} \times 255
\end{equation}
where $I_{i,j}$ indicates the value of the pixel in row $i$ and column $j$, $I_{min}$ indicates the minimum pixel value in $I_{hf}$ and $I_{max}$ indicates the maximum pixel value in $I_{hf}$.
\\\indent As shown in the second row of Fig. \ref{fig5}, the high-frequency image provides rich structural cues that are effectively exploited by the proposed FGSA model to enhance landmark detection accuracy. In addition, the proposed FMB-loss leverages complementary information from multiple datasets. Its effectiveness is validated in Fig. \ref{fig7}, where under challenging conditions such as occlusion and makeup, the highlighted regions demonstrate that the model achieves more accurate landmark localization when equipped with FMB-loss. 

\textbf{Analysis of different frequency-domain feature extraction schemes.} Since different frequency-domain extraction methods affect landmark localization accuracy, we further examine three schemes, including SRM\cite{han2021fighting}, Wavelet transform\cite{yuan2025feature} and FFT\cite{heckbert1995fourier}. As shown in Table \ref{frequency}, SRM achieves strong performance on 300W-Chall and WFLW, whereas Wavelet transform performs relatively better on AFLW but is less effective on 300W-Chall and WFLW, suggesting its limited ability to capture discriminative structural cues under challenging conditions. In contrast, our FFT-based approach achieves a favorable balance across all datasets, obtaining the best results on WFLW and COFW while maintaining competitive accuracy on 300W-Chall and AFLW. These findings demonstrate that FFT provides a more stable and effective means of capturing high-frequency facial structural information, thereby delivering superior or comparable performance across diverse benchmarks.

\textbf{Analysis of FMB-loss in comparison with other multi-task learning balancing losses. }DWA \cite{liu2019end} computes the importance of each task by taking the ratio of the current loss to the previous loss and then applying a smoothing operation. This importance weight is used to balance the convergence speeds of different tasks, encouraging the model to pay more uniform attention across tasks. Uncertainty Weighting (UW) \cite{kendall2018multi} introduces learnable parameters to model task-dependent uncertainty, thereby balancing differences in loss variances and mitigating the effect of task heteroscedasticity on the training process. As shown in Table \ref{mtl}, both DWA and UW yield only marginal improvements over the baseline across all evaluation metrics. This limited gain can be attributed to the inherent characteristics of facial landmark detection, where different datasets typically share a common subset of landmarks. In such cases, adjusting task-level weights alone is insufficient to achieve effective optimization. In contrast, the proposed FMB-loss addresses this issue by operating at the landmark level, which enables more precise and adaptive optimization across individual landmarks, resulting in substantially improved performance in multi-dataset training. Note that FGSA model isn't included in this experiment.
\begin{table}
	\caption{COMPUTATIONAL COMPLEXITY AND PARAMETERS COMPARISIONS WITH STATE-OF-THE-ART METHODS}
	\begin{center}
		\begin{tabular}{p{4.8cm}p{2cm}p{2cm}p{3.5cm}}
			\hline
			Method & Params(M) & FLOPs(G) & $NME_{io}$ on 300W  \\
			\hline
			LAB(CVPR18) \cite{wu2018look}&	12.29&	18.85&	3.49\\
			PIPNET-R101(IJCV21) \cite{jin2021pixel}&45.7&10.5&3.19\\
			DTLD(CVPR22)\cite{li2022towards}&13.3&2.5&2.96\\
			PicassoNet(TNNLS23)\cite{wen2022picassonet}&	1.96& 0.11&  3.58\\
			SCE-MAE(CVPR24)\cite{yin2024sce} &85.3&-&3.95\\
			\hline
			\textbf{FGTBT(ours)}&	\textbf{53.4}&	\textbf{103.45}&	\textbf{3.06}\\
			\hline
		\end{tabular}
	\end{center}
	\label{efficient}
\end{table}
\\\indent\textbf{Efficiency Analysis.} As shown in Table \ref{efficient}, although the proposed FGTBT model demonstrates competitive accuracy, its computational cost is relatively high, with 53.4M parameters and 103.45G FLOPs, making it less efficient compared to some lightweight CNN-based approaches. However, it outperforms Transformer-based methods such as SCE-MAE \cite{yin2024sce} by achieving a 0.89 improvement in 300W $NME_{io}$ with fewer parameters. This inefficiency mainly stems from adopting a Transformer architecture and the design objective of handling multiple datasets simultaneously within a unified framework. Unlike prior methods that require training separate models for different datasets, our approach only needs to maintain a single model, effectively reducing training and deployment overhead. Moreover, the end-to-end design enables the learning of robust and generalizable representations.
\\\indent\textbf{Comparison with Transformer-based Methods.} To further validate the effectiveness of the proposed method, we compare FGTBT with representative Transformer-based face landmark detection approaches, such as DTLD\cite{li2022towards} and SCE-MAE\cite{yin2024sce}. As shown in Table \ref{efficient}, FGTBT achieves a 0.89 improvement in  on the 300W dataset compared to SCE-MAE, while using fewer parameters. This demonstrates that our method not only leverages the Transformer architecture effectively but also achieves higher parameter efficiency. Regarding DTLD, the model requires a pretrained ResNet backbone and is trained specifically for a single dataset. In contrast, FGTBT is fully end-to-end and capable of handling multiple datasets within a unified framework. Although DTLD exhibits lower computational cost and competitive accuracy, its training pipeline is more complex and demands separate models for each dataset, which increases deployment overhead.
\\\indent\textbf{Limitations.} One drawback of our method is its relatively high inference cost: the model requires 103.45 GFLOPs, which limits its applicability to real-time scenarios. In addition, the performance may degrade on very low-resolution faces, since the model relies on sufficient visual details to extract informative frequency and geometry cues.
\subsection{Ablation study}

In this section, we conduct ablation experiments to validate the effectiveness of the components proposed in this work. Specifically, we evaluate the contributions of the backbone network, the Frequency-Guided Structure-Aware (FGSA) model and the Fine-grained Multi-task Balancing Loss (FMB-loss). Experiments are performed on the WFLW dataset and its expression subset. The results are presented in Table \ref{ablation}.

\textbf{Influence of FMB-loss.} The introduction of FMB-loss leads to noticeable performance gains on the WFLW dataset. This improvement stems from FMB-loss's capability to achieve a more precise and effective balance among multiple tasks. By assigning appropriate importance to underrepresented landmarks, the model is guided to attend to them more effectively, thereby enhancing alignment accuracy across datasets with diverse landmark definitions.

\textbf{Influence of the FGSA model.} The FGSA model improves performance on the WFLW-expression subset. This can be attributed to its ability to model facial structure information, which is particularly beneficial under conditions of facial expression variation. By guiding structural constraint learning with high-frequency information, the FGSA model effectively preserves crucial shape details, thereby enhancing the robustness and accuracy of facial landmark detection under expression variations.

\begin{table}
	\caption{ABLATION STUDY OF FGTBT ON WFLW BENCHMARK DATASET}
	\vspace{-1em}
	\begin{center}
		\begin{tabular}{p{6cm}p{2cm}p{3.6cm}}
			\hline
			Method & WFLW-full & WFLW-expression \\
			\hline
			FGTBT w/o FGSA \& FMB-loss &4.64&4.96	\\
			FGTBT w/o FGSA&4.47&4.74	\\
			\hline
			\textbf{FGTBT(ours)}&\textbf{4.42}&\textbf{4.54}	\\
			\hline
		\end{tabular}
	\end{center}
	\vspace{-1em}
	\label{ablation}
\end{table}

\section{Conclusion}

In this paper, we presented the Frequency-Guided Task-Balancing Transformer (FGTBT) for robust facial landmark detection under large pose variations, illumination changes, and diverse facial expressions. By jointly integrating the Fine-Grained Multi-Task Balance loss (FMB-loss) and the Frequency-Guided Structure-Aware (FGSA) module within a unified Transformer architecture, FGTBT effectively addresses the long-standing challenge of learning from multiple datasets with inconsistent annotation. Beyond performance gains across benchmarks, FGTBT offers a deployment-friendly unified modeling paradigm. Unlike conventional methods that rely on separate models for different datasets, FGTBT supports training and deployment with a single model across different datasets, reducing model maintenance costs.

Despite these advantages, the current framework still faces challenges in computational efficiency and robustness under low-resolution inputs. Future work will address these issues by incorporating causal inference to disentangle structural landmark cues from resolution-related noise, and by exploring spiking neural networks as an energy-efficient modeling alternative. These directions aim to improve efficiency while preserving structural sensitivity.

\leftline{ {\bf Acknowledgements}} This work is supported by the National Natural Science Foundation of China (Grant No.6257155 and 62002233) and the Natural Science Foundation of Hubei Province, China (Grant No. 2024AFB992).



\quad





\bibliographystyle{unsrt}
\bibliography{ref}

\end{document}